\documentclass[a4paper]{svproc}
\usepackage{url}

\usepackage{url}
\usepackage{svg}
\usepackage[backend=biber,
            hyperref=true,
            url=false,
            isbn=false,
            doi=false,
            backref=false,
            style=ieee,
            natbib=true,
            mincitenames=1,
            maxcitenames=1,
            citestyle=numeric-comp,
            sorting=nyt,
            block=none]{biblatex}
            
\usepackage{amsmath,amssymb}
\usepackage{graphicx}
\usepackage{soul}
\usepackage{adjustbox}
\usepackage{booktabs}
\usepackage{multirow}
\usepackage{verbatim}
\usepackage{xcolor} 
\usepackage{listings}
\usepackage{multicol}
\usepackage{diagbox}
\usepackage{subfig}
\usepackage{hyperref}
\usepackage{xspace}
\usepackage{textcomp}
\usepackage{gensymb}
\usepackage{marginnote}
\usepackage{tabularx}
\usepackage{wrapfig}
\usepackage{siunitx} 
\usepackage{caption}
\graphicspath{{figures/}}
\newcommand{\ba}{\mathbf{a}}

\newcommand{\bo}{\mathbf{o}}

\newcommand{\bxi}{\boldsymbol{\xi}}
\newcommand{\ie}{i.e.,\xspace}
\newcommand{\eg}{e.g.,\xspace}

\newcommand{\method}{GPT-Fabric\xspace}

\usepackage{pifont}
\usepackage{microtype}

\begin{document}
\mainmatter              
\title{GPT-Fabric: Smoothing and Folding Fabric\\ by Leveraging Pre-Trained Foundation Models}
\titlerunning{GPT-Fabric}  
%
\author{Vedant Raval$^*$ \and Enyu Zhao$^*$ \and Hejia Zhang \and Stefanos Nikolaidis \and Daniel Seita}
\authorrunning{Vedant Raval, Enyu Zhao, et al.} 
%
\tocauthor{Vedant Raval, Enyu Zhao, Hejia Zhang, Stefanos Nikolaidis, Daniel Seita}
\institute{University of Southern California, Los Angeles, CA 90089, USA, \\
\email{\{ravalv, enyuzhao, seita\}@usc.edu}
\quad {\footnotesize (* equal contribution)}
}
\maketitle              

\vspace{-1em}
\begin{abstract}
Fabric manipulation has applications in folding blankets, handling patient clothing, and protecting items with covers. It is challenging for robots to perform fabric manipulation since fabrics have infinite-dimensional configuration spaces, complex dynamics, and may be in folded or crumpled configurations with severe self-occlusions. 
Prior work on robotic fabric manipulation relies either on heavily engineered setups or learning-based approaches that create and train on robot-fabric interaction data. 
In this paper, we propose \method for the canonical tasks of fabric smoothing and folding, where GPT directly outputs an action informing a robot where to grasp and pull a fabric. 
We perform extensive experiments in simulation to test \method against prior methods for smoothing and folding.
\method matches the state-of-the-art in fabric smoothing, and also achieves comparable performance with most prior fabric folding methods tested, even without explicitly training on a fabric-specific dataset (\ie zero-shot manipulation). 
Furthermore, we apply \method in physical experiments over 10 smoothing and 12 folding rollouts. Our results suggest that \method is a promising approach for high-precision fabric manipulation tasks. 
Code, prompts, videos, {\color{black} and supplementary material} are available at 
\url{https://tinyurl.com/gptfab}.
\end{abstract}
\section{Introduction}\label{sec:intro}
Robot fabric manipulation has potential to address a wide variety of real-world applications, including dressing assistance~\cite{deep_dressing_2018,assistive_gym_2020}, folding and unfolding laundry~\cite{maitin2010cloth,unfolding_rf_2014}, and manufacturing textiles~\cite{Torgerson1987VisionGR}.
However, robot fabric manipulation is challenging due to the infinite-dimensional configuration space of fabrics and the complex dynamics resulting from robot-fabric interaction~\cite{manip_deformable_survey_2018,grasp_centered_survey_2019}, which hinder the use of traditional motion planning techniques. 
Thus, fabric manipulation remains an active area of research, of which recent works have used machine learning to train fabric smoothing~\cite{seita_fabrics_2020,lerrel_2020} and folding~\cite{fabricflownet,folding_fabric_fcn_2020}, with the intent of generalizing to different fabric configurations or targets. While these works show promising results, they require robot-fabric interaction data, either in simulation or in real. This data may consist of demonstrations~\cite{mo2022foldsformer}, or ``random'' interaction data~\cite{VCD_cloth}. 
This raises the question of whether it is possible to get similar performance without creating or training on a fabric-related dataset. 

\begin{figure}[t]
\center
\includegraphics[width=0.68\textwidth]{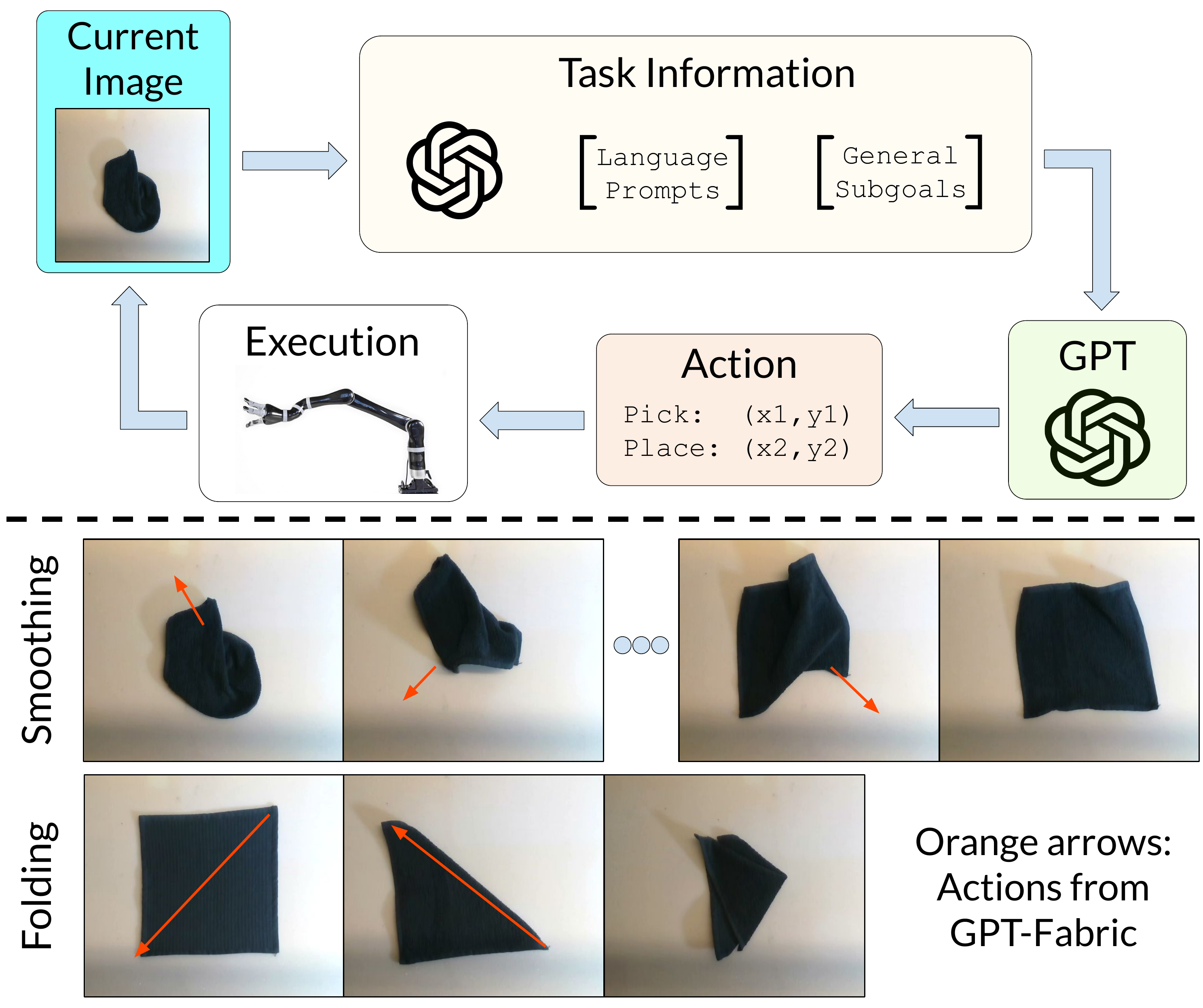}
\caption{
\textbf{Top}: high-level overview of \method. The input to the foundation model (GPT) is the current image observation of the fabric and the task information. The latter might include fabric manipulation strategies generated by a VLM, natural language descriptions to prompt the foundation models, and  (for folding tasks) the subgoal sequence targets (see Figure~\ref{fig:subgoals-sim}). \method directly produces actions (\eg pick-and-place) for a robot. \textbf{Bottom}: example rollouts of \method for smoothing and folding.
}
\vspace{-15pt}
\label{fig:pull}
\end{figure}

In parallel to these works on fabric manipulation, the AI research community has recently experienced an explosion of interest in foundation models~\cite{Bommasani2021FoundationModels} such as GPT-4~\cite{GPT-4} and Gemini~\cite{Gemini}. 
These models are trained on broad data and are capable of ``zero-shot'' generation of images, language, and code. 
In robotics, a standard way to employ foundation models is as a high-level planner. For example,~\cite{progprompt} and~\cite{codeaspolicies} use large language models to generate code describing high-level steps for a robot to execute. Consider when a robot must put food in a refrigerator; these approaches may produce code specifying: (1) go to kitchen, (2) open refrigerator, (3) go to counter, (4) get food, and (5) put in refrigerator, but this abstracts away low-level control. 
How, then, can we leverage the broad knowledge inherent in such foundation models for complex, low-level deformable object manipulation?

In this paper, we present \method (Figure~\ref{fig:pull}), that uses OpenAI's GPT to directly output low-level manipulation actions for robotic fabric manipulation. 
Despite the impressive performance of GPT on various tasks~\cite{GPT-4}, it is nontrivial to apply GPT to complex fabric manipulation tasks. 
As we later show in ablation experiments, we obtain poor fabric manipulation performance by naively providing the fabric image and prompting GPT. With \method, we show how to more effectively use GPT for fabric manipulation, and our method obtains comparable performance to most prior works and does not require creating a fabric-related dataset. 
More broadly, we hope this research helps facilitate using pre-trained foundation models for high-precision deformable object manipulation tasks. 

In summary, this paper contributes: 
\vspace{-0.8em}
\begin{enumerate}
    \item \method, a novel method for robot fabric manipulation that leverages GPT for low-level decision-making.
    \item State-of-the-art fabric smoothing and comparable fabric folding performance to most prior works, without needing to create a training dataset. 
    \item Ablation experiments investigating the importance of different aspects of \method.
\end{enumerate}

\section{Related Work}\label{sec:rw}


\subsection{Fabric Manipulation}

Fabric smoothing and folding are two canonical robotic fabric manipulation tasks. Pioneering research in this area has used bimanual robots with geometric algorithms. 
For example, a common approach to smooth fabrics was to leverage gravity to naturally smooth out the fabric to reveal corners, resulting in high success rates in smoothing crumpled towels~\cite{maitin2010cloth,unfolding_rf_2014}. 
Other researchers have studied how to grasp or smooth fabrics by utilizing geometric features, such as corner detection~\cite{willimon_2011} or fitting contours to clothing~\cite{laundry2012}. 
While effective, these prior approaches may require strict hardware, be time-consuming, or may have limited generalization to diverse fabric configurations. 

Deep learning in robotics has renewed interest in fabric manipulation, with the intent of leveraging powerful function approximators to learn complex motor skills from high dimensional observational data. 
Researchers have demonstrated fabric smoothing and folding using either imitation learning~\cite{hoque2022cloud,seita_fabrics_2020} or reinforcement learning~\cite{fabric_vsf_2020,lerrel_2020}, or both~\cite{DMFD_2022}. Other complementary techniques for data-driven fabric manipulation include latent space planning~\cite{latent_space_roadmap_2020,yan_fabrics_latent_2020}, learning dense visual descriptors~\cite{descriptors_fabrics_2021} and leveraging dynamic manipulation~\cite{xu2022dextairity,cloth_funnels}.  

Methods that use learning, however, require suitable fabric interaction data, which can be obtained via demonstrations~\cite{mo2022foldsformer}, simulators~\cite{corl2020softgym,DavidBM_2024}, or from real world trials~\cite{folding_fabric_fcn_2020}. 
A naturally related question is whether one can reduce the data requirement for learning fabric manipulation. Prior work has explored this by leveraging appropriate \emph{representations}. For example,~\cite{VCD_cloth} show that learning with a particle-based fabric representation is more sample-efficient than learning from images or a latent space for smoothing, and~\cite{fabricflownet} show a similar benefit for fabric folding by leveraging optical flow. 
Eliminating the need for goal observations specific to each fabric configuration,~\cite{mo2022foldsformer} improve fabric folding results by employing space-time attention in a Transformer architecture.
Nonetheless, these prior works still require creating and training on fabric-related interaction data. 
In contrast, we propose a novel approach to leverage the broad knowledge in GPT to attain competitive fabric manipulation performance which avoids the need to create fabric-related training data. 
Our method also uses natural language to prompt GPT to directly output low-level actions, distinguishing it from work that uses pre-trained language embeddings to describe folding targets~\cite{language_cond_graph_dyn_2024}.

\subsection{Foundation Models in Robotics}

The emergence of foundation models~\cite{Bommasani2021FoundationModels} such as GPT-4~\cite{GPT-4} and Gemini~\cite{Gemini}, has revolutionized the field of Artificial Intelligence. 
These models, which include Large Language Models (LLMs), are trained on massive data at scale, and can be deployed for many downstream applications. While these foundation models are often used for text-related applications such as natural language generation, researchers have recently employed them for real-world \emph{robotics} tasks; see~\cite{real_world_robot_FMs_2024,FMs_robotics_2023_survey_1,FMs_robotics_2023_survey_2} for representative surveys.  
One way to use foundation models is as a high-level planner, which can generate a sequence of steps that a robot should take to achieve an objective~\cite{huang2022language,driess2023palme}. Building on this, prior work has also leveraged foundation models to generate code to specify a robot's policy~\cite{progprompt,codeaspolicies,huang2023voxposer}.
In addition, these models can generate reward functions for robots~\cite{ma2023eureka,yu2023language}, potentially by querying pairs of images of an agent's observations~\cite{RL-VLM-F}.  

Along with using foundation models for high-level tasks, there has been some recent work using them to specify low-level actions for decision-making. For example,~\cite{Walk_Robot_LLMs} demonstrate how to prompt GPT-4 to produce joint angles to make a quadruped robot walk.~\cite{GPT-Driver} and~\cite{mao2023agentdriver} use GPT models by recasting autonomous driving as a language modeling problem. 
Finally, among the most relevant recent works,~\cite{LLMs_zero_shot_traj} use GPT as a zero-shot planner to generate a dense sequence of end-effector robot poses, and benchmark this on a variety of language-conditioned tasks. 
Taking inspiration from these, we study the novel application of applying foundation models for complex, low-level deformable
object manipulation. 

\vspace{-5pt}
\section{Problem Statement}\label{sec:PS}

In this paper, we study quasi-static fabric smoothing and folding with a single-arm robot. We assume that there is one piece of fabric on a flat workspace. 
We use $\bxi_t$ and $\bo_t$ to represent, respectively, the fabric \emph{state} and \emph{image observation} at time $t$ in an interaction \emph{rollout}, which lasts up to a user-specified value of at-most $T$ time steps. 
A rollout transforms the initial fabric state $\bxi_0$ and image $\bo_0$ to the final state $\bxi_T$ and image $\bo_T$, respectively. 
More concretely, $\bxi_t \in \mathbb{R}^{N \times 3}$ is the set of $N$ particles (or ``points'') that form the fabric. Each particle has a 3D world coordinate position. 
In addition, $\bo_t \in \mathbb{R}^{H \times W \times c}$ is an RGB-D image with height $H$, width $W$, and $c=4$ channels (three for color, one for depth).

We specify a robot's actions with picking and placing poses: $\ba_t = \{x_{\rm pick}, x_{\rm place}\}$,
where the robot grasps the fabric at $x_{\rm pick}$, lifts the fabric by a small amount, drags it parallel to the workspace plane towards $x_{\rm place}$, where it then releases the fabric. Here, $x_{\rm pick}$ and $x_{\rm place}$ are 3D world coordinate positions. They may be derived from 2D image pixels $p_{\rm pick}$ and $p_{\rm place}$ via robot-camera calibration. This type of action primitive is standard in prior work on (quasi-static) robot fabric manipulation~\cite{fabric_vsf_2020,seita_fabrics_2020,lerrel_2020,VCD_cloth}.

Finally, we assume access to a corner detector which is \emph{not} fabric-specific (\eg Harris~\cite{harris_1988} or Shi-Tomasi~\cite{shi_tomasi_1994}). 

\begin{figure}[t]
\center
\includegraphics[width=0.85\textwidth]{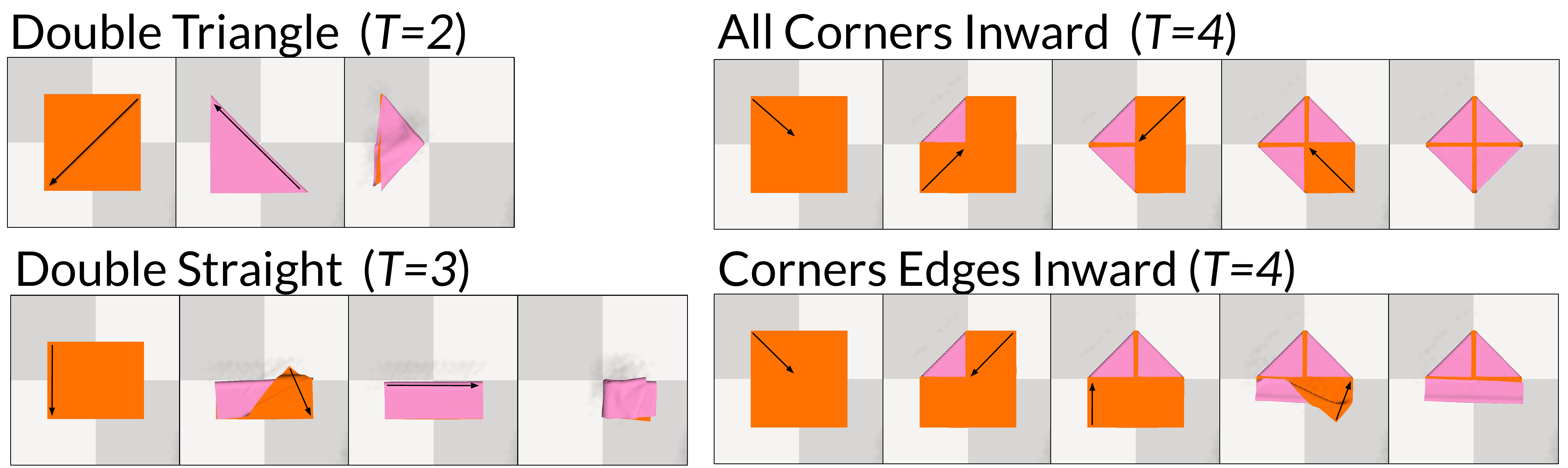} 
\caption{
Subgoal sequences for the four folding tasks we consider (from~\cite{mo2022foldsformer}) with maximum rollout lengths $T$. We use these for folding in simulation (Section~\ref{ssec:folding-exps}) and real (Section~\ref{sec:experiments_physical}).
}
\vspace*{-10pt}
\label{fig:subgoals-sim}
\end{figure}

Given this formulation, we consider the following tasks.

\vspace{-5pt}
\subsubsection{Smoothing}

The fabric starts in a crumpled configuration. We assume a priori that the objective is smoothing. 
As in prior work on smoothing~\cite{seita_fabrics_2020,VCD_cloth}, the objective is to improve \emph{fabric coverage} $C(\bxi_t) \in [0,1]$, which is the fraction of the fabric's area when projected on the 2D workspace $xy$-plane, compared to the area of a fully flat fabric. 
We specify $x_{\rm pick}$ as one of the $N$ points of the fabric state $\bxi_t$. We consider policies that specify the change in the $x$ and $y$ coordinates after $x_{\rm pick}$, which enables us to obtain $x_{\rm place}$. 
For experiments, we test maximum rollout lengths of $T \in \{5, 10, 20\}$ for consistency with~\cite{VCD_cloth}.

\vspace{-10pt}
\subsubsection{Folding}

The fabric starts flat on the workspace. {\color{black} As in prior work~\cite{fabricflownet,mo2022foldsformer}, we assume access to a sequence of $T$ subgoal image observations $\{ \bo^{(g)}_1,\bo^{(g)}_2,\ldots, \bo^{(g)}_T\}$ representing the process to obtain the final desired fabric configuration for each folding task, since a single goal observation may not describe the full fabric state due to self occlusion. Going from each subgoal to the next is possible with a single pick-and-place action, however we do not assume access to this ground-truth action. Although a natural language description for each folding step could benefit a VLM-based method like ours, we do not assume access to this information to remain consistent with prior work.}

{\color{black} Following~\cite{mo2022foldsformer}, we use a generic demonstration subgoal sequence for each task used for different starting fabric configurations (see Figure~\ref{fig:subgoals-sim}). We add an arrow-based action visualization to aid our method in visual reasoning. Note that these arrows do not represent the ground-truth action at test time.}

We assume that at test time, the robot must execute a novel sequence of subgoals, and we input $\bo_{t}^{(g)}$ and $\bo_{t+1}^{(g)}$ to the robot to reach the fabric state $\bxi_{t+1}$. We represent the pick $p_{\rm pick}$ and place $p_{\rm place}$ points as 2D image coordinates in the image $\bo_t$ of the fabric. 

\microtypecontext{expansion=sloppy}{In simulation, we evaluate using \emph{mean particle distance error} between the achieved and goal fabrics~\cite{mo2022foldsformer,fabricflownet}. In real world, without ground-truth particle information, we evaluate by human inspection as done in prior work~\cite{descriptors_fabrics_2021,folding_fabric_fcn_2020,fabric_vsf_2020}.}
 
\vspace{-5pt}
\section{Method}\label{sec:method}


\subsection{\method: Overall Structure}\label{ssec:overal}

\begin{figure*}[t]
\center
\includegraphics[width=\textwidth]{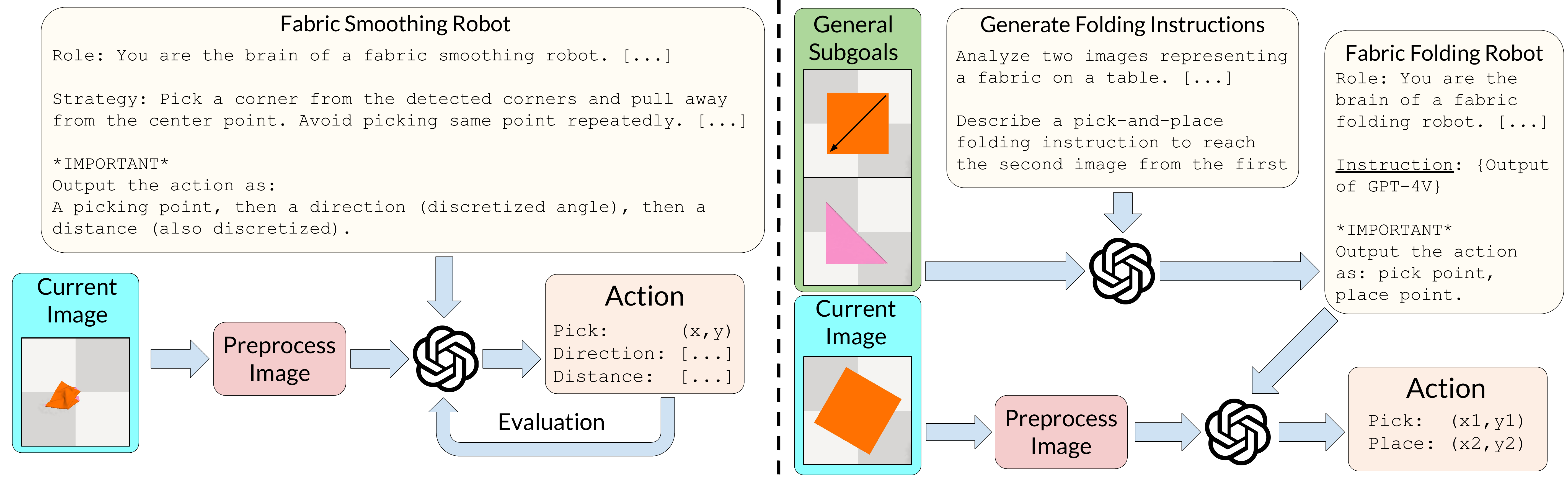} 
\caption{
Our \method method, for smoothing (left) and folding (right). The input prompt to GPT includes the current RGB-D image of the fabric $\bo_t$ (only showing RGB above) and the task information. For folding, this information includes subgoal sequences (see Figure~\ref{fig:subgoals-sim}). We also preprocess $\bo_t$ for \method and use an \emph{evaluation module} for smoothing; see Figure~\ref{fig:gpt-smoothing} for details.}
\vspace*{-12pt} 
\label{fig:system}
\end{figure*}

As suggested in prior work~\cite{seita_fabrics_2020,VCD_cloth,fabric_vsf_2020,fabricflownet}, the corners and center of a fabric are the most relevant for a successful fabric manipulation. 
Thus, for each time $t$, we preprocess the image observation $\bo_t$ to extract these keypoints to be part of the prompt to GPT. 
While the exact preprocessing of $\bo_t$ and prompting are different for smoothing and folding (see Sections~\ref{ssec:smoothing} and~\ref{ssec:folding}), both use depth to mask out the background, and off-the-shelf corner detectors (as per our assumptions in Section~\ref{sec:PS}) to obtain estimates of the fabric corners. These detectors are not fully reliable and could cause errors. This processed information is part of the prompts. If the prompt does not include in-context examples, then we refer to this method as \textbf{GPT-Fabric (zero-shot)}. 

Given the prompt, GPT produces a pick pose as a 2D pixel $p_{\rm pick}$ in the image representation $\bo_t$, which we convert to $x_{\rm pick}$. GPT also produces a place pose as a 2D pixel $p_{\rm place}$ for folding or as a tuple of the moving direction and the moving pixel distance for smoothing. In both cases, we convert this to $x_{\rm place}$. The robot, either in simulation or in real, can utilize $x_{\rm pick}$ and $x_{\rm place}$ to perform the action. 

\vspace{-5pt}
\subsection{\method for Smoothing}\label{ssec:smoothing}

\begin{figure}[t]
\center
\includegraphics[width=0.85\textwidth]{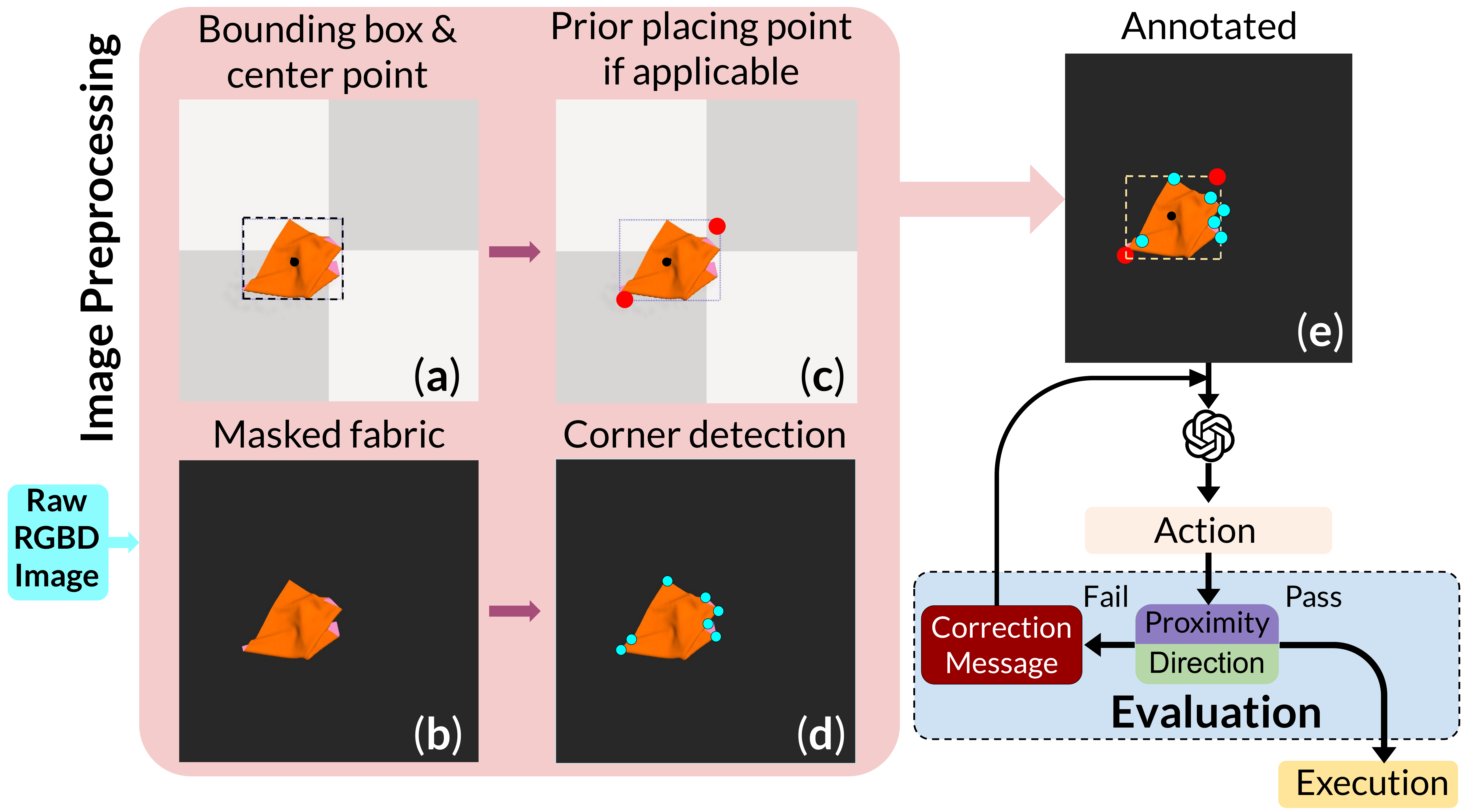} 
\caption{
Details of image annotation and evaluation for smoothing (see Section~\ref{ssec:smoothing} for more). From RGB-D image $\bo_t$, we get a bounding box and approximate fabric center (a) via masking (b). If applicable, we annotate the \emph{prior} placing point and its ``symmetric'' point about the fabric center (c). We detect corners (d) on the masked image, then combine (c) and (d) to get image (e) as input to GPT-4V. We use an evaluation module to verify GPT's output. If it fails, we ask GPT to try again with a correction message. 
}
\vspace*{-18pt}
\label{fig:gpt-smoothing}
\end{figure}

We use Shi-Tomasi Corner Detection~\cite{shi_tomasi_1994} to detect possible fabric corners, and approximate the center of the fabric via its bounding box. 
Annotation can improve GPT's reasoning about images~\cite{yang2023setofmark}, so we annotate the RGB image with the detected corners, bounding box, and approximate fabric center. For all time steps after the initial step, we also annotate the RGB image with the prior action's \emph{placing point} location. Figure~\ref{fig:gpt-smoothing} visualizes the above preprocessing and annotation procedure for smoothing. After obtaining the annotated image, we combine it with a natural language prompt for the foundation model GPT-4V, as visualized in Figure~\ref{fig:system} (left). 
The language prompt describes the smoothing task, an explanation of the annotated RGB image, and a high-level strategy. The prompt decomposes the strategy into low-level instructions to guide GPT-4V to:

\begin{itemize}
    \item pick one of the detected corners as the picking point. 
    \item pick a pull orientation angle that moves ``outwards'' from the fabric's center. 
    \item pick the largest pull distance which avoids dragging the fabric center too far from the image center. 
\end{itemize}

{\color{black}We discretize pull orientations and distances to simplify the task and request a structured output for the action.} The following is a summary of the prompts:
\smallskip
\color{gray}
\begin{lstlisting}
|- Your task is to provide a pick and place action with the following information:|
|1. The pick location. Pick one of the given corners \{...\}. \textcolor{green}{\# GPT-4V's answer here}|
|2. The pull angle. Pick one of \{0,pi/4, pi/2,..., 7*pi/4\}. \textcolor{green}{\# GPT-4V's answer here}|
|3. The pull length. Pick one \{0.1, 0.25, 0.5, 0.75, 1.0\},where 1 is the length of|
|\phantom{xxx}a flattened fabric edge. \textcolor{green}{\# GPT-4V's answer here}|
\end{lstlisting}
\color{black}
\smallskip

Given the proposed action, we use an \emph{evaluation module} to verify if the action follows our high-level strategy (see Figure~\ref{fig:gpt-smoothing}). This module checks the picking point and its Euclidean distance to the prior picking point, as well as the orientation. We execute the action if it satisfies our tests, and otherwise query GPT again for another proposed action. We limit to three queries per time step. {\color{black} Instead of filtering detected corners by distance and orientation, the evaluation module guides GPT to learn from its mistakes and improve smoothing, avoiding over-engineering by not directly supplying the optimal corners.}

\subsection{\method for Folding}\label{ssec:folding}


At each time $t$, we use GPT-4V to analyze the current subgoals $\bo_{t}^{(g)}$ and $\bo_{t+1}^{(g)}$ and provide a natural language folding instruction. We also preprocess the RGB-D image $\bo_t$ to obtain a set of candidate corners via Harris Corner Detection~\cite{harris_1988} and use the fabric mask to estimate the fabric center. This generated instruction, along with the detected key-points, is passed to the foundation model (either GPT-3.5 or GPT-4) to return a pick-and-place policy to achieve the desired subgoal. We construct a system prompt which offers a universal context related to fabric folding, inspired by~\cite{GPT-Driver}. By doing this, we do \emph{not} hand-design a lengthy prompt that is specific to a folding task. Instead, GPT-4V produces the instructions for each type of fold. If we want to achieve a new folding task, it should be possible to use \method simply by specifying the folding subgoal sequence.
The following is a summary of the prompts:

\smallskip
\color{gray}
\begin{lstlisting}
|- Given two folding subgoal images with an arrow representing a pick-and-place action \{...\},| 
|provide an instruction to achieve the transition between the images. \textcolor{green}{\# GPT-4V's answer here}|
|- Given the above folding instruction, provide a pair of pick and place points on the cloth|
|from the corners \{...\} to achieve this folding step. \textcolor{green}{\# GPT-4 or GPT-3.5's answer here}|
\end{lstlisting}
\color{black}
\smallskip

This system does not require expert demonstrations and can act as a zero-shot manipulator. However, we can leverage the in-context learning capabilities of GPT by providing demonstrations while generating folding instructions as well as while getting the pick and place points. Figure~\ref{fig:system} (right) visualizes the folding pipeline. If we use in-context learning, we call this method \textbf{GPT-Fabric (in-context)}. Given the limitations of the fine-tuning API to support only text, we leave a \emph{\method (fine-tuned)} extension to future work.

\section{Simulation Experiments}\label{sec:sim-exps}

For consistent comparisons with prior works, we leverage SoftGym simulation~\cite{corl2020softgym} to test \method. 

\subsection{Fabric Smoothing Experiments}\label{ssec:smoothing-exps}

We compare \method for single-arm, quasi-static fabric smoothing with data and results from Visible Connectivity Dynamics (VCD)~\cite{VCD_cloth}. Lin~et~al.~\cite{VCD_cloth} showed that VCD attains superior coverage performance compared to Fabric-VSF~\cite{fabric_vsf_2020}, Contrastive Forward Model~\cite{yan_fabrics_latent_2020}, and Maximal Value under Placing~\cite{lerrel_2020}.
Thus, we use the same square fabric in the 40 starting, crumpled configurations from~\cite{VCD_cloth} for testing \method on smoothing. To compare results, we directly use statistics reported in~\cite{VCD_cloth}. In addition, we also download and deploy their publicly released trained model weights on the same fabric configurations. 
We perform one rollout for zero-shot \method for each of the 40 starting fabric configurations and each of the three maximum possible allowed actions (5, 10, and 20), for a total of 120 rollouts.

We evaluate using the Normalized Improvement (NI) in coverage. NI computes the increased covered area normalized by the maximum possible improvement, ${NI=\frac{s-s_0}{s_{max}-s_0}}$, where $s_0$, $s$, and $s_{max}$ are the initial, achieved, and maximum possible fabric coverage.
As in~\cite{VCD_cloth}, smoothing rollouts end early when NI exceeds 0.95. 
In our quantitative smoothing results, to be consistent with~\cite{VCD_cloth}, we report the 25$\%$, 50$\%$, and 75$\%$ percentiles $(Q_{25},Q_{50},Q_{75})$ of NI; the number before $\pm$ is $Q_{50}$ and the number after $\pm$ is $\max(|Q_{50}-Q_{25}|,|Q_{75}-Q_{50}|)$. 

\subsection{Fabric Folding Experiments}\label{ssec:folding-exps}

We primarily compare \method for single-arm, quasi-static fabric folding with data and results from Foldsformer~\cite{mo2022foldsformer}. Mo~et~al.~\cite{mo2022foldsformer} performed controlled experiments for fabric folding, and showed that Foldsformer achieves the lowest (\ie best) mean particle distance error compared to FabricFlowNet~\cite{fabricflownet}, Fabric-VSF~\cite{fabric_vsf_2020}, and Lee~et~al.~\cite{folding_fabric_fcn_2020}. 
Thus, we leverage the open-source Foldsformer data for our experiments, and we directly use results that they report for comparisons. 
We test on their same starting fabric configurations, their same (test-time) subgoal sequences for folding, and the same folding types (see Figure~\ref{fig:subgoals-sim}). 
For \emph{Double Triangle}, \emph{All Corners Inward}, and \emph{Corners Edges Inward}, we use a square fabric in 40 different configurations per task, combining 10 distinct side lengths, ranging from 31.25cm to 36.875cm, and 4 rotation angles of 0\degree, 30\degree, 45\degree, and 60\degree. For \emph{Double Straight}, we use a rectangular fabric with the 10 sizes ranging from 31.25cm$\times$27.5cm to 36.875cm$\times$32.5cm and the above mentioned rotation angles.

We set the temperature of GPT-4V to 0.2 to allow some diversity in the generated instructions. 
We consider both GPT-4 and GPT-3.5 to reason from these instructions while performing experiments for zero-shot as well as in-context version. For in-context learning, we generate demonstrations (see Section~\ref{ssec:data-sizes}) by leveraging a script provided by Foldsformer~\cite{mo2022foldsformer}. 
Since the GPT models are not deterministic, we run \method for five trials for each fabric starting state and subgoal sequence. We average results from those five to get metrics for a single starting state and target combination. We then average over all those combinations to obtain our final mean particle distance error results.  To be consistent with~\cite{mo2022foldsformer}, for our quantitative results, the number before $\pm$ is the mean and the number after is the standard deviation.

\vspace{-10pt}
\subsection{Data Sizes}\label{ssec:data-sizes}


One of our motivations is to leverage the broad knowledge in GPT to avoid explicitly creating a dataset for fabric manipulation. We measure a method's ``data size'' by the amount of pick-and-place actions in its training data, where the only requirement is that the actions are applied on fabric. We do not distinguish between an action specific to a fabric manipulation task (\eg specific to smoothing, or to a \emph{Double Triangle} fold) or a ``random'' pick-and-place action on a fabric. These count the same towards the overall data size.

For smoothing, VCD~\cite{VCD_cloth} trained on 2000 pick-and-place actions. This is orders of magnitude smaller than Fabric-VSF~\cite{fabric_vsf_2020} trained on 106,725 actions, Contrastive Forward Module~\cite{yan_fabrics_latent_2020} trained on 400,000 actions, and Maximal Value of Placing~\cite{lerrel_2020}, trained on 250,000 actions. Lin~et~al.~\cite{VCD_cloth} benchmarked these prior methods using data sizes similar to those reported in the original papers. 

For folding, FabricFlowNet~\cite{fabricflownet} trained on 20,000 actions in simulation, Fabric-VSF~\cite{fabric_vsf_2020} used 106,725 actions in simulation, and Lee~et~al.~\cite{folding_fabric_fcn_2020} only used real world data consisting of 300 actions. 
Mo~et~al.~\cite{mo2022foldsformer} benchmarked these prior methods and their proposed Foldsformer method on a common dataset of 48,000 random actions in simulation. In addition, they used the same 100 task-specific demonstrations for all methods, adding more 200-400 actions per task. 


In contrast, \method (zero-shot) does not require creating a fabric-related interaction data. Moreover, we only consider 10 expert demonstrations per each action in each folding task for the in-context version of \method, adding 20 actions for Double Triangle, 30 actions for Double Straight, and 40 actions each for All Corners Inward and Corners Edges Inward, totalling to only 130 actions.

\begin{table*}[t]
  \setlength\tabcolsep{5.0pt}
  \centering
  \footnotesize
  \begin{adjustbox}{width=1\textwidth,center}
    \begin{tabular}{|l|rrr|c|}
    \toprule
        \multirow{1}{*}{Method} &
             \# of actions: 05  &  
             \# of actions: 10  &  
             \# of actions: 20  & 
        \multirow{1}{*}{Data Size} \\
        \midrule
        VCD$^\dagger$                  &  0.624 $\pm$ 0.217 &  0.778 $\pm$ 0.222 &  0.968 $\pm$ 0.307  & 2K \\
        VCD$^\S$                  &  0.472 $\pm$ 0.205 &  0.625 $\pm$ 0.222 &  0.791 $\pm$ 0.281  & 2K \\
        VCD, graph imitation$^\dagger$ &  0.692 $\pm$ 0.258 &  0.919 $\pm$ 0.377 &  $0.990 \pm 0.122$  & 2K \\
        Fabric-VSF$^\dagger$ & 0.321 $\pm$ 0.112 & 0.561 $\pm$ 0.127 & 0.767 $\pm$ 0.134 & $\sim$105K \\
        CFM$^\dagger$ & 0.053 $\pm$ 0.051 & 0.077 $\pm$ 0.053 & 0.109 $\pm$ 0.066 &  400K \\
        MVP$^\dagger$ & 0.399 $\pm$ 0.210 & 0.435 $\pm$ 0.137  & 0.421 $\pm$ 0.361 &  250K \\
        \midrule
        \method (zero-shot) &   0.733 $\pm$ 0.229 &  0.959 $\pm$ 0.240 & 0.986 $\pm$ 0.035& 0 \\
    \bottomrule
    \multicolumn{5}{l}{$^\dagger$\footnotesize{Reported by VCD authors}. $^{\S}$\footnotesize{Using trained model weights provided by VCD authors.}} \\
  \end{tabular}
  \end{adjustbox}
  \vspace{5pt}
  \caption{
  Results for fabric smoothing in simulation, comparing \method versus prior works on the same starting crumpled (square) fabric configurations from~\cite{VCD_cloth}. We report $Q_{50}\pm\max(|Q_{50}-Q_{25}|,|Q_{75}-Q_{50}|)$ for the normalized improvement in coverage; see Section~\ref{ssec:smoothing-exps} for details. This is \emph{not} the same statistic as we use in Table~\ref{tab:folding}. 
  } 
  \vspace*{-15pt}
  \label{tab:smoothing}
\end{table*}  

\begin{table*}[t]
  \setlength\tabcolsep{3.3pt}
  \centering
  \footnotesize
  \begin{adjustbox}{width=1.00\textwidth,center}
    \begin{tabular}{|l|cccc|c|}
    \toprule
        Method & 
        Double &  
        Double &  
        All Corners &
        Corners Edges &
        Data \\
        & 
        Triangle  &  
        Straight &  
        Inward  &
        Inward &
        Size \\
    \midrule
    Foldsformer$^\dagger$ & 19.64 $\pm$ 17.08 & 59.09 $\pm$ 44.82 & 3.06 $\pm$ 1.83 & 8.11 $\pm$ 7.96 & $>$48K \\
    Foldsformer$^\S$ & 14.00 $\pm$ 15.39 & 28.79 $\pm$ 30.65 & 2.94 $\pm$ 1.74 &  8.49 $\pm$ 2.20 & $>$48K \\
    FabricFlowNet$^\dagger$ & 102.20 $\pm$ 19.39 & 85.34 $\pm$ 19.67 & 33.64 $\pm$ 11.76 & 36.95 $\pm$ 9.95 & $>$48K \\  
    Lee et al.$^\dagger$ & 109.82 $\pm$ 39.96 & 114.71 $\pm$ 26.06 & 27.21 $\pm$ 11.47 & 70.67 $\pm$ 22.24 & $>$48K \\
    Fabric-VSF$^\dagger$ & 114.62 $\pm$ 35.46 & 116.94 $\pm$ 24.52 & 46.05 $\pm$ 23.38 & 51.82 $\pm$ 16.65 & $>$48K \\
    \midrule
    \method (GPT-4, zs) & 82.61 $\pm$ 6.74 & 76.63 $\pm$ 8.40 & 63.51 $\pm$ 12.84 & 79.94 $\pm$ 22.51 & 0 \\
    \method (GPT-3.5, zs) & 84.76 $\pm$ 8.51 & 73.24 $\pm$ 7.51 & 67.70 $\pm$ 12.31 & 98.99 $\pm$ 25.28 & 0\\
    \midrule
    \method (GPT-4, ic) & 43.43 $\pm$ 17.44 & 72.56 $\pm$ 13.57 & 57.53 $\pm$ 20.27 & 121.98 $\pm$ 15.44 & 130 \\
    \method (GPT-3.5, ic) & 51.89 $\pm$ 18.55 & 69.24 $\pm$ 9.68 & 61.20 $\pm$ 18.93 & 128.57 $\pm$ 22.25 & 130 \\
    \bottomrule
    \multicolumn{6}{l}{$^\dagger$\footnotesize{Reported by Foldsformer authors}. $^{\S}$\footnotesize{Using trained model weights provided by Foldsformer authors}.}
    \end{tabular}
  \end{adjustbox}
  \vspace{5pt}
  \caption{
  Results for fabric folding in simulation, comparing \method versus prior works on the same folding subgoal targets from Mo~et~al.~\cite{mo2022foldsformer}. We report the mean particle distance error (mm) for the four fold types, and the number after $\pm$ is the standard deviation. For \method, ``zs'' is ``zero-shot'' and ``ic'' is ``in-context.''
  }
  \vspace*{-15pt}
  \label{tab:folding}
\end{table*}

\vspace{-5pt}
\section{Simulation Results}\label{sec:sim-results}

\begin{figure*}[t]
\center
\includegraphics[width=1.00\textwidth]{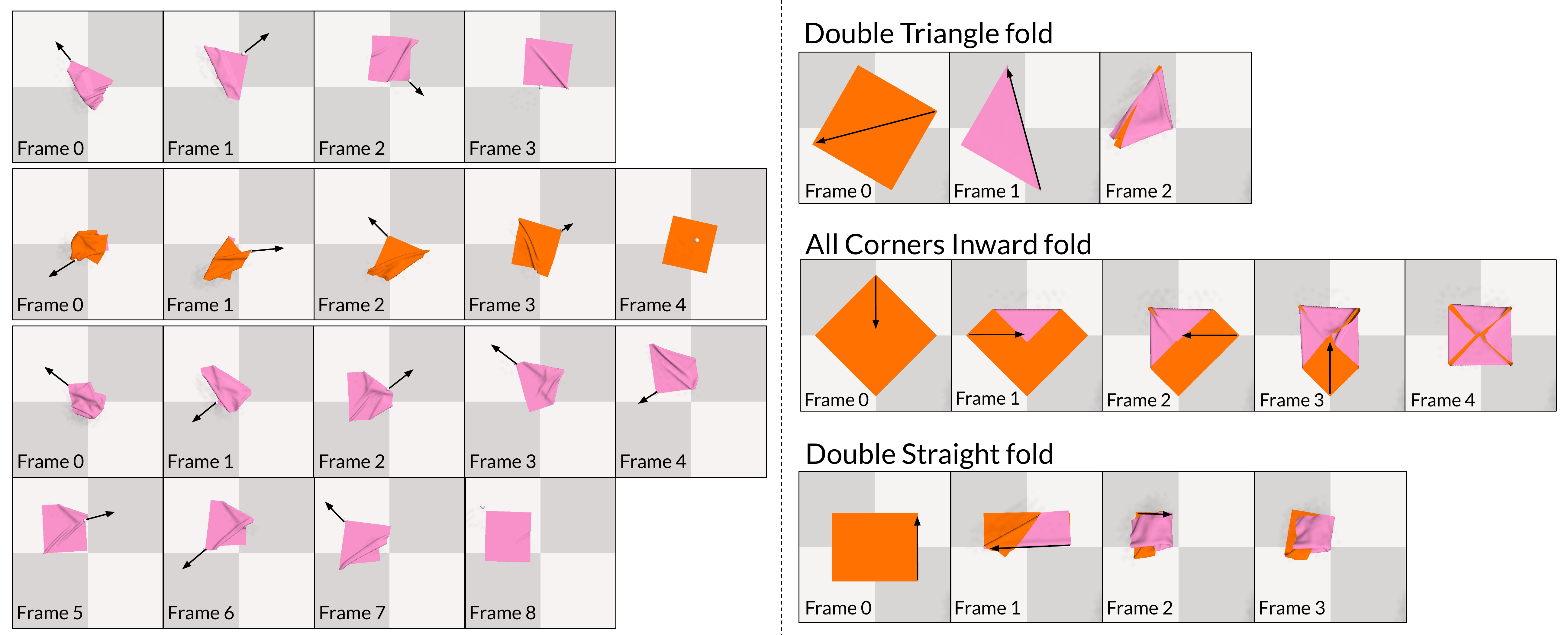} 
\caption{
Qualitative results for smoothing (left) and folding (right) from \method in SoftGym simulation. We show three smoothing rollouts of varying frame lengths, where each frame shows one pick-and-place action. In all three smoothing rollouts, \method achieved NI$>$0.95. We show three examples of folding rollouts with different folding subgoals (see Figure~\ref{fig:subgoals-sim}). \method was unable to achieve qualitatively good results for \emph{Corners Edges Inward}.
}
\vspace*{-15pt}
\label{fig:results-sim}
\end{figure*}

\subsection{Fabric Smoothing Results}\label{ssec:smoothing-results-sim}

Figure~\ref{fig:results-sim} (left) shows qualitative results of zero-shot \method for smoothing a square fabric in simulation. 
Table~\ref{tab:smoothing} reports the normalized improvement (NI) in coverage, for different total allowed pick-and-place actions, using numbers directly from Supplementary Table 4 in~\cite{VCD_cloth} for all baselines. 
Due to space limitations, we defer additional metrics to {\color{black} the supplementary material.}

The results show \method (zero-shot) matches or outperforms the best baseline (VCD with graph imitation), with a higher median for 5-action and 10-action horizons, while VCD has a higher median NI for 20-action horizon. {\color{black} Since \method (zero-shot) outperforms the baselines, we do not include an in-context version for smoothing.}

\begin{table}
\vspace{10pt}
\centering
\footnotesize
\begin{adjustbox}{width=0.7\textwidth,center}
\begin{tabular}{|c|c|}
\toprule
Method & NI \\
\midrule
Random discretized action & 0.203 $\pm$ 0.230 \\ 
No image preprocessing \& no eval. module & 0.021 $\pm$ 0.144 \\ 
No eval. module                           & 0.540 $\pm$ 0.220 \\ 
No verifying pick point proximity in eval. module        & 0.682 $\pm$ 0.222 \\
\midrule
\method (zero-shot)                     & 0.733 $\pm$ 0.229 \\
\bottomrule
\end{tabular}
\end{adjustbox}
\vspace{5pt}
\caption{Normalized improvement (NI) for \method ablations for smoothing in simulation after 5 pick-and-place actions. The bottom row is the full \method method (reported in Table~\ref{tab:smoothing}). }
\vspace*{-15pt}
\label{tab:ablation_smoothing}
\end{table}

\vspace{-9pt}
\subsubsection*{Ablation Studies}

We ablate components of \method and report the NI in smoothing in Table~\ref{tab:ablation_smoothing}, using the same notation as in Table~\ref{tab:smoothing}. 
We test four ablations: (1) random selection of action parameters, (2) no image preprocessing of $\bo_t$, (3) no evaluation (``eval.'') module to verify GPT's output, or (4) only using part of it by removing the proximity check (see Figure~\ref{fig:gpt-smoothing}). 
The results indicate that removing components of \method leads to worse performance. See {\color{black} the supplementary material} for more details.

\begin{figure*}[t]
\center
\includegraphics[width=1.00\textwidth]{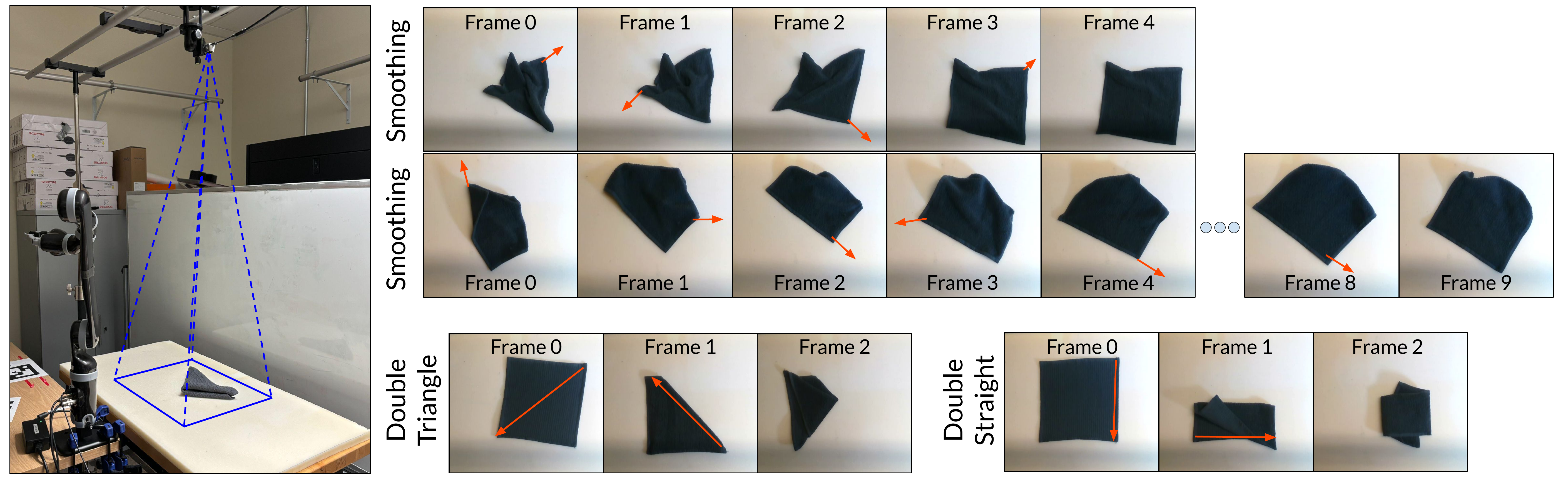}
\caption{
\textbf{Left}: physical setup for fabric manipulation experiments. The Kinova Jaco is next to the flat workspace with blue lines indicating the top-down camera view from the Intel RealSense camera. \textbf{Right}: two smoothing rollouts and two folding rollouts of different targets (\emph{Double Triangle} and \emph{Double Straight}). While both folds achieve reasonable final performance, \method only uses two actions for the \emph{Double Straight} fold, skipping an intermediate step shown in the subgoals (see Figure~\ref{fig:subgoals-sim}). 
}
\vspace*{-13pt}
\label{fig:physical}
\end{figure*}

\vspace{-5pt}
\subsection{Fabric Folding Results}\label{ssec:folding-results-sim}

Figure~\ref{fig:results-sim} (right) shows qualitative results of \method for fabric folding in simulation. 
We report results in Table~\ref{tab:folding}, where the values from baseline methods are directly from Table I in Mo~et~al.~\cite{mo2022foldsformer}. Overall, the results suggest that \method attains competitive performance compared to prior works of FabricFlowNet and Fabric-VSF, even as a zero-shot manipulator. While results are worse than Foldsformer, we use a fraction of the data (in the in-context version) and we do not need to train a model on fabric manipulation data. 

We observe that for \emph{Double Triangle} and \emph{Double Straight}, which are relatively harder tasks for FabricFlowNet and Fabric-VSF, \method can obtain better performance even in the zero-shot setting. Furthermore, performing in-context learning with either GPT-4 or GPT-3.5 generally lowers the mean particle distance errors. In particular, using just 20 demonstrations lowered the mean particle distance error to almost half for the \emph{Double Triangle} fold. 

{\color{black} On the other hand, Table~\ref{tab:folding} suggests that \emph{Corners Edges Inward} and \emph{All Corners Inward} folds are challenging for \method, likely due to implicit biases against the fabric center in our tests. We perform prompt ablation studies for \method, demonstrating how different prompt components influence folding performance, including the impact of certain biases. We defer the details to the supplementary material. Please refer to the project website for the generated folding instructions.}





Overall, compared to FabricFlowNet, \method obtains competitive, if not better, mean particle distance error. However, \method only performs single-arm folding, while FabricFlowNet elegantly supports bimanual manipulation, which is not tested in our experiments or in Mo~et~al.~\cite{mo2022foldsformer}. We leave a ``bimanual \method'' extension to future work.

\vspace{-5pt}
\section{Physical Experiments}\label{sec:experiments_physical}

\vspace{-5pt}
We perform physical experiments with \method (zero-shot) for smoothing and \method (in-context) for folding. This does \emph{not} require additional real world data. Compared to simulation, we adjust how we detect fabric keypoints (\eg to account for noisy depth cameras) and a camera-robot calibration procedure to convert pixels to ``world positions.'' 
See the {\color{black} supplementary material} for details.
We set up a workstation with a 1-inch thick foam which has dimension \SI{60}{\centi\meter} by \SI{105}{\centi\meter}. We mount an Intel RealSense d435i RGBD camera at a height of about \SI{1}{\meter} (see Figure~\ref{fig:physical}). We use a 6-DOF Kinova Jaco robot manipulator with a KG-3 gripper, where we remove one of the fingers to make it similar to a standard parallel-jaw gripper as used in prior work on fabric manipulation. We use a square \SI{30.5}{\centi\meter} by \SI{30.5}{\centi\meter} grey dish cloth, which is similar to the fabric sizes in~\cite{fabricflownet,mo2022foldsformer}.

\subsection{Experiment Protocol}


For smoothing and folding, a human initializes a crumpled fabric and a flat fabric, respectively. We perform 10 rollouts for smoothing and 12 rollouts for folding, three times for each of the four folding targets.
Unlike prior work~\cite{fabricflownet, mo2022foldsformer}, we use subgoal images from simulation instead of real-world fabric. Each action involves the robot going to a ``pre-pick'' pose \SI{5}{\milli\meter} above the target, then lowering and grasping. Then, the robot lifts by \SI{5}{\milli\meter}, moves to the ``pre-place'' pose, then lowers and releases its grip. 
The maximum number of actions per rollout is either 10 (for smoothing) or is dependent on the number of subgoals (for folding).

\vspace{-3pt}
\subsection{Physical Results}

We show qualitative rollouts of our method for smoothing and folding in Figures~\ref{fig:pull} and~\ref{fig:physical}. The results suggest that \method can achieve qualitatively acceptable fabric smoothing and folding results, despite how we do not directly train it on a fabric-related dataset. 
For smoothing, we obtain $0.806\pm0.295$ NI over the 10 rollouts, of which 6 also got at least 0.85 coverage. For folding, \method gets a 2/3 success rate for Double Triangle, and a 2/3 success rate for Double Straight. It does not succeed on the other two fold types since GPT frequently fails to set the fabric's center as the placing point{\color{black}, consistent with our discussion in Section~\ref{ssec:folding-results-sim}. A real-world folding episode is considered successful if the system achieves the desired fabric configuration within the maximum number of actions.}

\begin{wrapfigure}{r}{0.48\textwidth}
\vspace*{-31pt}
\center
\includegraphics[width=0.46\textwidth]{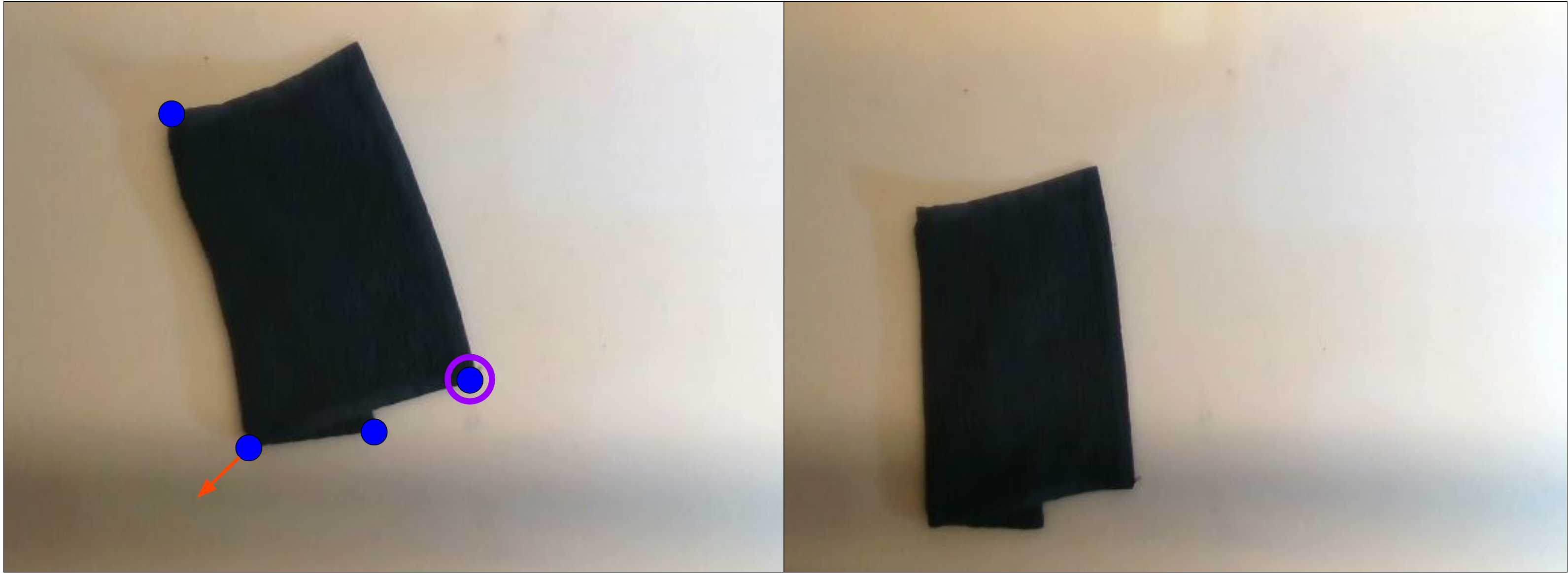}
\caption{
An example inefficiency of \method from GPT. Among detected corners (blue circles), \method chose to pull at the bottom left corner. This barely adjusts the coverage, in contrast to potentially pulling the purple bordered point. 
}
\vspace*{-20pt}
\label{fig:failure}
\end{wrapfigure}

We categorize failures in three cases: (1) hardware limits (including robot-camera calibration or robot imprecision), (2) corner detection errors, or (3) \method limitations due to GPT's incorrect reasoning. For instance, the robot can be off by \SI{1}{\centi\meter} even after careful calibration or may fail to reach picking points due to range limits. In instances where fabric corners are tucked underneath, the corner detection of \method may fail to detect them.
Finally, even with accurate corner detection, GPT may suggest suboptimal picking points (\eg see Figure~\ref{fig:failure}). 




%
%

\vspace{-3pt}
\section{Conclusion}\label{sec:conclusion}


We present \method, a system for using GPT for low-level fabric manipulation. Our experiments show that \method can attain competitive or better results versus most prior methods, without explicitly training on a fabric-related dataset. 
In future work, we will extend \method to other deformable object manipulation tasks. 
We hope this work motivates research on using GPT, as well as other current and future foundation models, for low-level motor control skills for complex robotic manipulation tasks.

\printbibliography

\newpage
\appendix
\chapter*{Supplementary Material for GPT-Fabric:
Smoothing and Folding Fabric by Leveraging
Pre-Trained Foundation Models}
\section{Additional Prompt Details}\label{appendix:prompt}

This section includes a description of the reasoning and intuition behind the development of the prompts and the overall structure of the \method pipeline. The full prompts can be found on the project website\footnote{\url{https://tinyurl.com/gptfab}}.

\subsection{\method for Smoothing}


The design of our high-level smoothing strategy incorporates human intuition about smoothing. We believe that repeatedly choosing the same corner to pick can drag the fabric out of the camera's view, as well as picking and placing opposite corners back-to-back, which simply moves the fabric back and forth. When the picking point is chosen, an efficient direction to pull it can be the vector starting from the center point of the fabric and pointing towards the chosen picking point.

We also employ a verification procedure, visualized in Figure~\ref{fig:gpt-smoothing} of the main text, to align the output of GPT-4V with the high-level strategy. We first perform a \emph{proximity check} to ensure that the chosen picking point is sufficiently far from the last step's picking point (or its symmetric point). If this passes, we then conduct a \emph{direction check} to verify whether the predicted moving direction significantly deviates from the actual direction which originates from the center point and extends to the picking point. We take the prediction of GPT-4V as the final output if both tests pass. Otherwise, we provide GPT-4V with a correction message explaining why the previous prediction fails, and the visualization of the output action to the original input and let it infer again. If GPT-4V's output fails to pass for three consecutive times, we discard the picking point's proximity check and only perform the direction check. 

If the detected corners are all near the prior step's picking point and its symmetric point, this can make it impossible to pass the proximity check. However, the direction check is essential, as a ``wrong'' direction can be catastrophic. In our cases, GPT-4V occasionally makes the mistake of choosing the exact opposite direction, pulling the chosen picking point towards the center point.
Therefore, we still conduct the direction check, believing that pulling the chosen picking point away from the center point will not worsen the situation as it will likely stretch the fabric to the pulling direction.



\subsection{\method for Folding} \label{appendix:folding-prompt}
Being consistent with prior work, we assume access to just the folding subgoal images, without any textual details corresponding to the folding task. Since \method is dependent on the vision reasoning abilities of GPT-4V for generating the folding instructions, we require that the generated instructions are reasonably accurate for the downstream LLM to obtain the pick and place actions effectively.

In preliminary experiments, we observed that the model exhibited errors ranging from disregarding all the information in the subgoal images by interpreting them as placeholder thumbnails to being biased towards suggesting diagonal folds or center folds in the generated folding instructions. To address these issues, we construct a language prompt with five major components:
\begin{enumerate}
    \item \emph{Visual Context}: A brief explanation of the context represented by the subgoal images to ensure that GPT-4V treats them as valid image input.
    \item \emph{Pick-and-place Arrow}: A description of the pick-and-place action facilitated by a black arrow attached to the subgoal images depicting a folding step.
    \item \emph{Center Pivoting}: A reasoning logic that utilizes the action description specified above to estimate the possible place point location relative to the fabric center to ensure that the place point is not biased towards being one of the fabric corners.
    \item \emph{Instruction Constraint}: An explicit requirement to generate the folding instruction in terms of the pick and place point to ensure that the downstream LLM could utilize that to return an executable action.
    \item \emph{Output Format}: An explicit output format specification that could be parsed to get an instruction to be used by the downstream LLM.
\end{enumerate}

See Section~\ref{appendix:folding-ablation} for our experiments on ablating these five prompt components. 

For the downstream LLM (either GPT-4 or GPT-3.5) to effectively utilize this folding instruction, we construct an additional natural language prompt to describe the task, ensuring that the model is aware of the purpose of the context while generating an output that follows a fixed format in order to parse an executable action. We use a chain-of-thought prompting mechanism to have the LLM produce the reasoning behind the generated pick-and-place action. 

\section{Additional Experiment Details}

\subsection{Simulation Experiments: Ablations for Smoothing}
\label{app:smoothing-ablation}

The main text discusses our ablations with GPT-Fabric for smoothing, along with the experimental results. The ablations are:
\begin{itemize}
    \item \emph{Random discretized action:} We conduct the corner detection to the raw image observation $\bo_t$ and randomly choose one from those detected corners as the picking point $x_{pick}$. We also randomly choose the pulling action's direction and distance from the same discretized list in the prompt (see Section~\ref{ssec:smoothing} in the main text). This is a way to compare how much GPT outperforms random guessing. 

    \item \emph{No image preprocessing \& no eval. module:} We only conduct the corner detection to the raw image observation $\bo_t$ and provide the detected corners' coordinates to GPT; we adapt the system and user prompt to the removal of the image preprocessing module.
    We also remove the evaluation module and directly execute the action from GPT. 

    \item \emph{No eval. module:} We remove the evaluation module. 

    \item \emph{No verifying pick point proximity in eval. module:} In the evaluation module, we only perform the \emph{direction check} to verify GPT's output action's direction and execute the action if it passes. 
\end{itemize}

Our results in the main text show that removing any major component in \method for leads to significantly worse smoothing performance. 
The performance of \emph{No image preprocessing \& no eval. module} is worse than \emph{Random discretized action}, suggesting that fabric smoothing is an extremely challenging task for GPT alone, underlining the importance of our image preprocessing and evaluation modules. We also witness worse performance in \emph{No eval. module} where GPT occasionally pulls the picking point towards the center point, resulting in an enormous coverage decrease. 

\begin{table}
\centering
\footnotesize
\begin{tabular}{|c|c|c|}
\toprule
Method & Double Triangle & All Corners Inward \\
\midrule
No Visual Context & $58.88 \pm 21.44$ & $95.75 \pm 22.07$\\ 
No Pick-and-place Arrow & $97.53 \pm 11.49$ & $72.79 \pm 20.06$\\ 
No Center Pivoting & $65.29 \pm 19.05$ & $148.57 \pm 28.60$\\
No Instruction Constraint & $58.88 \pm 24.92$ & $68.42 \pm 16.76$\\
No Output Format & $87.85 \pm 18.15$ & $36.09 \pm 14.55$\\
\midrule
\method (zero-shot) & $68.52 \pm 20.42$ & $75.59 \pm 23.11$ \\
\bottomrule
\end{tabular}
\vspace*{5pt}
\caption{Mean particle distance errors (mm) for \method prompt ablations for folding in simulation. Lower numbers are better. The last row refers to our full method.}
\label{tab:ablation_folding}
\end{table}

\vspace{-30pt}
\subsection{Simulation Experiments: Ablations for Folding} \label{appendix:folding-ablation}

We ablate five core components of the prompt (see Section~\ref{appendix:folding-prompt}). We perform these ablations for \emph{Double Triangle} and \emph{All Corners Inward} folding using a similar experimental setup as discussed in the main text. For these experiments, we average over three trials of \method (zero-shot) and use GPT-4 as the downstream LLM to generate executable pick and place points. The results are in Table~\ref{tab:ablation_folding}.

Our results show that removing any major component leads to a performance degradation in at least one folding type in general, highlighting the importance of all five components. This benefit could be due to biases in \method predictions in the absence of certain components. For instance, on ablating the \emph{Output Format} component, we observe that \method often chooses the fabric center as the placing point for the folding actions, leading to a significantly lower mean particle distance error for \emph{All Corners Inward} fold but a higher mean particle distance error for \emph{Double Triangle} fold when compared with \method (zero-shot). The results for \method (zero-shot) in Table~\ref{tab:ablation_folding} differ from the results for \method (GPT-4, zero-shot) in the Table~\ref{tab:folding} of the main text, due to different number of trials and since GPT is nondeterministic even with temperature of 0. 

\subsection{Simulation Experiments: Additional Results}
We report the normalized coverage in Table~\ref{tab:smoothing-extended}. This is an alternative metric (also used in~\cite{VCD_cloth}) compared to normalized improvement, which we report in the main text. These metrics under consideration are evaluated on the same set of experiments.


\begin{table*}
  \setlength\tabcolsep{5.0pt}
  \centering
  \footnotesize
    \begin{adjustbox}{width=1.00\textwidth,center}
        \begin{tabular}{|l|rrr|c|}
        \toprule
            \multirow{1}{*}{Method} &
                 \# of actions: 05  &  
                 \# of actions: 10  &  
                 \# of actions: 20  & 
            \multirow{1}{*}{Data Size} \\
            \midrule
            VCD$^\dagger$                  &  0.776 $\pm$ 0.132 &  0.872 $\pm$ 0.128 &  0.985 $\pm$ 0.187  & 2K \\
            VCD$^\S$                  &  0.643 $\pm$ 0.121 &  0.745 $\pm$ 0.101 &  0.876 $\pm$ 0.151  & 2K \\
            VCD, graph imitation$^\dagger$ &  0.837 $\pm$ 0.150 &  0.966 $\pm$ 0.236 &  0.996 $\pm$ 0.076  & 2K \\
            \midrule
            \method (zero-shot) &   0.887 $\pm$ 0.120 &  0.980 $\pm$ 0.112 & 0.992 $\pm$ 0.036 & 0 \\
        \bottomrule
        \multicolumn{5}{l}{$^\dagger$\footnotesize{Reported by VCD authors}. $^{\S}$\footnotesize{Using trained model weights provided by VCD authors.}} \\
      \end{tabular}  
    \end{adjustbox}
  \vspace{5pt}
  \caption{
  Extended results for fabric smoothing in simulation, comparing \method versus prior works on the same starting crumpled (square) fabric configurations from Lin~et~al.~\cite{VCD_cloth}. 
  We format the table in a similar manner as in the main text, except that we report \emph{normalized coverage} instead of \emph{normalized improvement}, and we only compare \method with VCD variants. 
  } 
  \vspace*{-10pt}
  \label{tab:smoothing-extended}
\end{table*} 

\vspace{-20pt}
\subsection{Real World Experiment Setup}

As mentioned in the main text, we incorporate some changes in our system, specifically for preprocessing the input image, when compared with simulation. Limited by the accuracy of the depth camera in the real world, we do not utilize depth information for assistance and instead use the top-down RGB image directly. We use a contour-finding algorithm to obtain the contour of the fabric and approximate the achieved coverage $s$ by its area. We construct a bounding box around the contour and use its center point to approximate the center point of the fabric. We apply Harris Corner Detector~\cite{harris_1988} for folding tasks and Shi-Tomasi Corner Detector~\cite{shi_tomasi_1994} for smoothing tasks to the top-down RGB image, retaining only the detected corners within the bounding box as candidate picking points. We use a standard calibration procedure to convert the ``camera position'' to the ``world position'' for the robot in both tasks, converting the picking and placing points computed by \method to real world positions, where the robot can execute its actions.





\subsection{Inference Time}
Using our setup of two NVIDIA RTX 4090 GPUs, the average inference time for planning one pick-and-place smoothing action with \method in simulation is approximately \SI{18.01}{\second}, which is slower compared to \SI{12.70}{\second} for VCD (as reported by the authors using their setup of four NVIDIA RTX 2080Ti GPUs). Similarly, the average inference time for planning one pick-and-place folding action with \method is approximately \SI{12.46}{\second}, which is slower compared to \SI{0.02}{\second} for Foldsformer~\cite{mo2022foldsformer} when replicated on our setup. These numbers are partially due to querying the OpenAI API multiple times in both workflows. In future work we will explore reducing the number of API queries.

\subsection{Budget}

Running experiments in this paper involved frequent calls to the OpenAI API to use GPT models. In all, we paid approximately 2300 USD for the API, which includes the preliminary trials and all the reported experiments in simulation and in real.

\end{document}